\def\year{2022}\relax
\documentclass[letterpaper]{article} % DO NOT CHANGE THIS
\usepackage{aaai21}  % DO NOT CHANGE THIS
\usepackage{times}  % DO NOT CHANGE THIS
\usepackage{helvet} % DO NOT CHANGE THIS
\usepackage{courier}  % DO NOT CHANGE THIS
\usepackage[hyphens]{url}  % DO NOT CHANGE THIS
\usepackage{graphicx} % DO NOT CHANGE THIS
\usepackage{natbib}  % DO NOT CHANGE THIS AND DO NOT ADD ANY OPTIONS TO IT
\usepackage{caption} % DO NOT CHANGE THIS AND DO NOT ADD ANY OPTIONS TO IT
\usepackage{amsmath}
\usepackage{amssymb}
\usepackage{algorithm}
\usepackage{algorithmic}
\usepackage{multirow}
\usepackage{multicol}
\usepackage{color}
\usepackage{bm}
\newcommand{\tabincell}[2]
{\begin{tabular}
{@{}#1@{}}#2\end{tabular}}

\title{Unsupervised Temporal Video Grounding with Deep Semantic Clustering}
\author{Daizong Liu\textsuperscript{\rm 1,2$^\dagger$}, Xiaoye Qu\textsuperscript{\rm 2$^\dagger$}, Yinzhen Wang\textsuperscript{\rm 3$^\dagger$}, Xing Di\textsuperscript{\rm 4}, Kai Zou\textsuperscript{\rm 4}, Yu Cheng\textsuperscript{\rm 5}, \\ Zichuan Xu\textsuperscript{\rm 6}, Pan Zhou\textsuperscript{\rm 1*} \\}
\affiliations{
\textsuperscript{\rm 1}The Hubei Engineering Research Center on Big Data Security, School of
Cyber Science and Engineering, \\ Huazhong University of Science and Technology\\
\textsuperscript{\rm 2}School of Electronic Information and Communication,
Huazhong University of Science and Technology\\
\textsuperscript{\rm 3}School of Computer Science and Technology, 
Huazhong University of Science and Technology\\
\textsuperscript{\rm 4}ProtagoLabs Inc \
\textsuperscript{\rm 5}Microsoft Research \
\textsuperscript{\rm 6}Dalian University of Technology
\\ \{dzliu, xiaoye, yinzhenwang, panzhou\}@hust.edu.cn, \{xing.di, kz\}@protagolabs.com, \\ yu.cheng@microsoft.com,  z.xu@dlut.edu.cn
}
\begin{document}

\maketitle

\begin{abstract}
Temporal video grounding (TVG) aims to localize a target segment in a video according to a given sentence query. Though respectable works have made decent achievements in this task, they severely rely on abundant video-query paired data, which is expensive and time-consuming to collect in real-world scenarios. In this paper, we explore whether a video grounding model can be learned without any paired annotations. To the best of our knowledge, this paper is the first work trying to address TVG in an unsupervised setting. Considering there is no paired supervision,
we propose a novel \textbf{D}eep \textbf{S}emantic \textbf{C}lustering \textbf{N}etwork (DSCNet) to leverage all semantic information from the whole query set to compose the possible activity in each video for grounding. Specifically, we first develop a language semantic mining module, which extracts implicit semantic features from the whole query set. Then, these language semantic features serve as the guidance to compose the activity in video via a video-based semantic aggregation module. Finally, we utilize a foreground attention branch to filter out the redundant background activities and refine the grounding results.
To validate the effectiveness of our DSCNet, we conduct experiments on both ActivityNet Captions and Charades-STA datasets. The results demonstrate that DSCNet achieves competitive performance, and even outperforms most weakly-supervised approaches. 
\end{abstract}

\section{Introduction}
\begin{figure}
\begin{center}
{\includegraphics[width=0.48\textwidth]{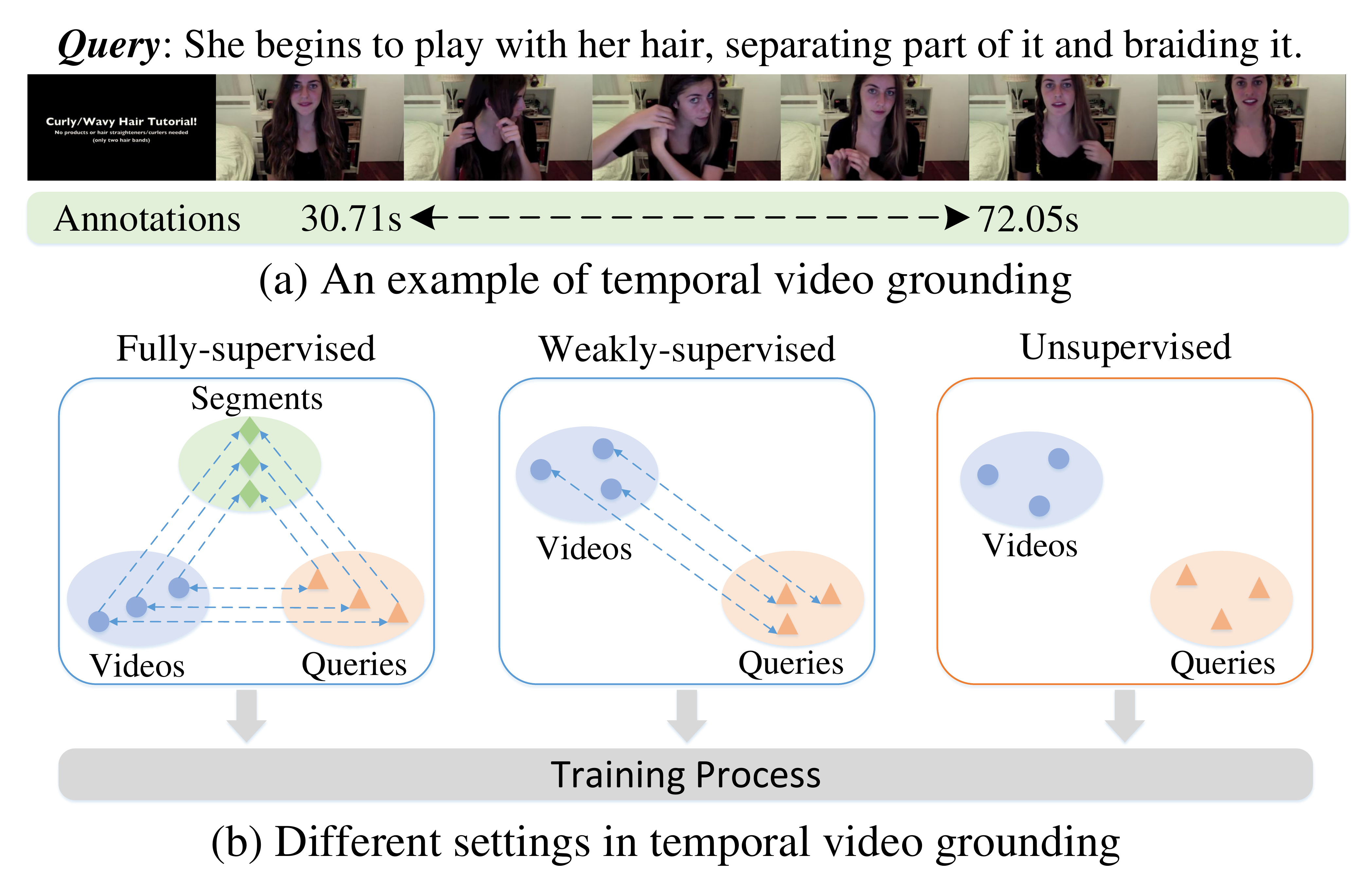}}
\end{center}
\vspace{-10pt}
\caption{(a) An example of temporal video grounding. (b) Different from the fully-supervised (paired video-query and detailed segment annotations) and weakly-supervised (only paired video-query knowledge) settings in TVG, there is no supervised information in unsupervised setting.}
\vspace{-10pt}
\label{fig:introduction}
\end{figure}

Temporal video grounding (TVG) is an important yet challenging task in video understanding, which has drawn increasing attention due to its vast potential applications, such as activity detection \cite{zhao2017temporal} and human-computer interaction \cite{singha2018dynamic}. As depicted in Figure \ref{fig:introduction} (a), it aims to localize a segment in a video according to the semantic of a sentence query. 
% In particular, it requires predicting the accurate start and end timestamps of the segment.

Most previous methods \cite{wang2020temporally,liu2021context,zhang2019man,liu2020jointly,Alpher10,chen2020rethinking} proposed for this task are under fully-supervised setting. Some of them \cite{liu2021context,Alpher33,liu2020jointly} match the pre-defined segment proposals with the query and then select the best candidate. Others \cite{Alpher08,Alpher11,Alpher12,chen2020rethinking} directly predict the temporal boundary of the video segment. However, these methods are data-hungry, requiring a large amount of fully annotated data. For instance, the widely used ActivityNet Captions dataset contains 20,000 videos and 100,000 matched sentence queries with their corresponding segment boundaries. Manually annotating such a huge amount of data is very time-consuming and labor-intensive. 
To alleviate this problem, recent works explore a weakly-supervised setting \cite{Alpher14,Alpher15,Alpher17,zhang2020regularized} where only paired videos and queries are available in the training stage. Though they leave out the segment annotations in training, these methods still need to access the abundant knowledge of the matched video-query pairs. 

In this paper, we focus on how to learn a video grounding model without any supervision, which excludes both paired video-query knowledge and corresponding segment annotations, as shown in Figure \ref{fig:introduction} (b). 
% For this unsupervised task, a straightforward way is to utilize a two-stage method, which first selects the most relevant video among the video set for each sentence, and then localizes the target segment in the selected video.
% However, both models for ranking videos and localizing segments are infeasible to achieve satisfying performance due to high model complexity and the lack of annotated data.
Considering there is no annotated information, what we can access to is only the internal information in the queries and videos.  
As different words or phrases in different queries may share potentially similar semantic, 
% instead of directly ranking and matching the videos to a certain query, 
we can mine all deep semantic representations from the whole query set, and then compose the possible activities in each video according to these language semantic for further grounding.
Therefore, the crucial and challenging part of our work lies in how to capture the deep semantic features of the queries and how to aggregate different semantics for composing the contents of the target segments.

\begin{figure}[t!]
\begin{center}
%\fbox{\rule{0pt}{1.8in} \rule{0.9\linewidth}{0pt}}
\includegraphics[width=0.48\textwidth]{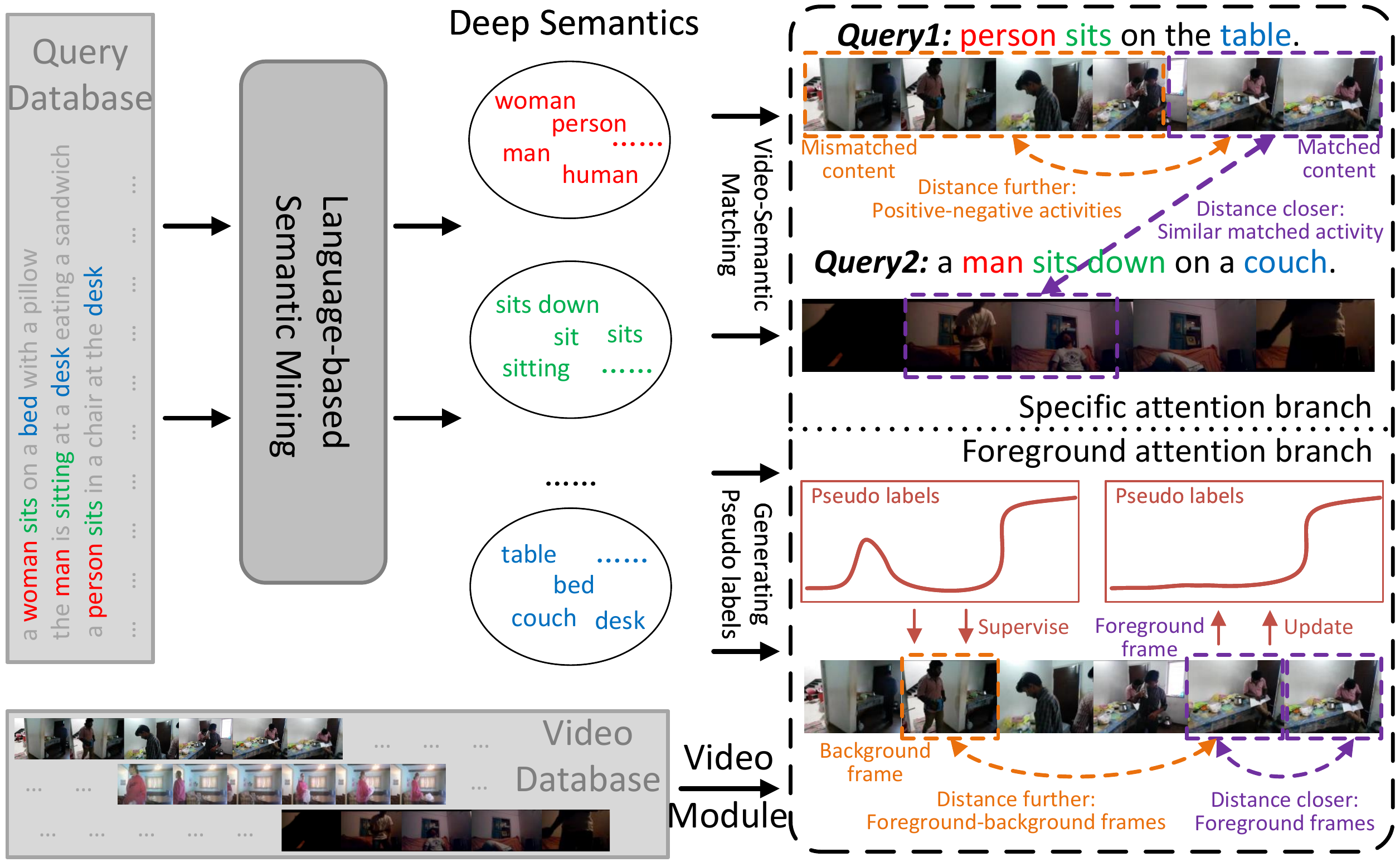}
\end{center}
\vspace{-10pt}
\caption{The main idea of our proposed method, where we only show several semantic clusters for example.}
\vspace{-10pt}
\label{fig:description}
\end{figure}

% As shown in Figure \ref{fig:description}
To this end, we propose a novel approach to solve this problem, called \textbf{D}eep \textbf{S}emantic \textbf{C}lustering \textbf{N}etwork (DSCNet), which mines the deep semantic features from the query set to compose possible activities in each video. Specifically, we first leverage an encoder-decoder model to build a language-based semantic mining module for query encoding, where the learned hidden states are taken as the extracted deep semantic features. In particular, we collect such semantic features from the whole query set and then cluster them to different semantic centers, where features of similar meanings are adjacent. Subsequently, a video-based semantic aggregation module, containing a specific attention branch and a foreground attention branch, is further developed to compose corresponding activities guided by the extracted deep semantic features.
% as shown in Figure \ref{fig:description}. 
% The specific attention branch aggregates different semantics for matching and composing the contents of the activity segments. To further filter out the background information, the foreground attention branch is designed to filter out the background activities composed by the irrelevant semantic features. 
For the specific attention branch, it aggregates different semantic for matching and composing the contents of the activity segments. We utilize this branch to generate better video representations by ensuring that the composed activities containing the same semantic have closer distance than the dissimilar ones, and the positive-negative activity in the same video have large distance. To further filter out the background information in each video, the foreground attention branch is designed to distinguish the foreground frames. 
% which filters out the most relevant activities from redundant semantic features. 
The details of our main grounding process is shown in Figure \ref{fig:description}.
During the training stage, we utilize the pseudo labels, which are obtained from the deep semantic features, as guidance to refine the video grounding model with an iterative learning procedure.
To sum up, our main contributions are as follows:
\begin{itemize}
% \vspace{-3pt}
\item To the best of our knowledge, this is the first work to address temporal video grounding in the unsupervised setting. Without supervision, we solve the task with the proposed DSCNet, which learns to compose the activity contents guided by the deep language semantic.
% \vspace{-8pt}
\item 
% We utilize an auto-encoder model to learn the sentence reconstruction, and take the learned hidden states as the deep semantic clues of each sentence. 
We use an encoder-decoder module to obtain the semantic features for all queries and divide them into different clusters to represent different semantic meanings. 
Then we propose a two-branch video module, where specific attention branch aggregates the query semantic to match the segment, and the foreground attention branch is utilized to distinguish the foreground-background activities.
% \vspace{-8pt}
\item We conduct comprehensive experiments on the ActivityNet Captions and Charade-STA datasets. The results demonstrate the effectiveness of our proposed method, where DSCNet achieves decent results and outperforms most weakly-supervised methods.
\end{itemize}

\section{Related Work}
\noindent \textbf{Fully-supervised temporal video grounding.}
Most of the existing methods refer to fully-supervised setting where all video-sentence pairs are labeled in details, including corresponding segment boundaries.
Therefore, the main challenge in such setting is how to align multi-modal features well to predict precise boundary.
% \cite{Alpher06} and \cite{Alpher07} use sliding windows with various lengths to generate segment proposals to match the query.
% Instead of using the sliding windows, 
Some works \cite{qu2020fine,liu2021progressively,liu2021adaptive,liu2020reasoning,liu2022exploring,liu2022memory} integrate sentence information with each fine-grained video clip unit, and predict the scores of candidate segments by gradually merging the fusion feature sequence over time. Although these methods achieve good performances, they severely rely on the quality of the proposals and are time-consuming.
Without using proposals, the latest methods \cite{nan2021interventional,Alpher11,Alpher12} are proposed to leverage the interaction between video and sentence to predict the starting and ending frames directly. These methods are more efficient than the proposal-based ones, but achieve lower performance.

\noindent \textbf{Weakly-supervised temporal video grounding.}
% The aforementioned fully-supervised methods heavily rely on the datasets that require numerous manually labelled temporal annotations. 
To ease the human labelling efforts, several works \cite{Alpher14,Alpher15,Alpher16,Alpher17,zhang2020regularized,ma2020vlanet,tan2021logan} consider a weakly-supervised setting which only access the information of matched video-query pairs without accurate segment boundaries. \cite{Alpher15}
utilize the dependency between video and sentence as the supervision while
abandon the temporal ordered information. Their text-guided attention provides scores for segment proposals.
\cite{Alpher16} put forward a
module to reconstruct sentences and a proposal reward is based on the loss calculated using the target sentence and reconstructed one. 
Though these weakly-supervised methods do not rely on the temporal annotations, they still need the dependency between video and sentence as supervision. Different from them, we are the first to attempt to solve this task with an unsupervised approach that does not require any video-query dependency.

\begin{figure*}
% \vspace{-10pt}
\begin{center}
{\includegraphics[scale=0.40]{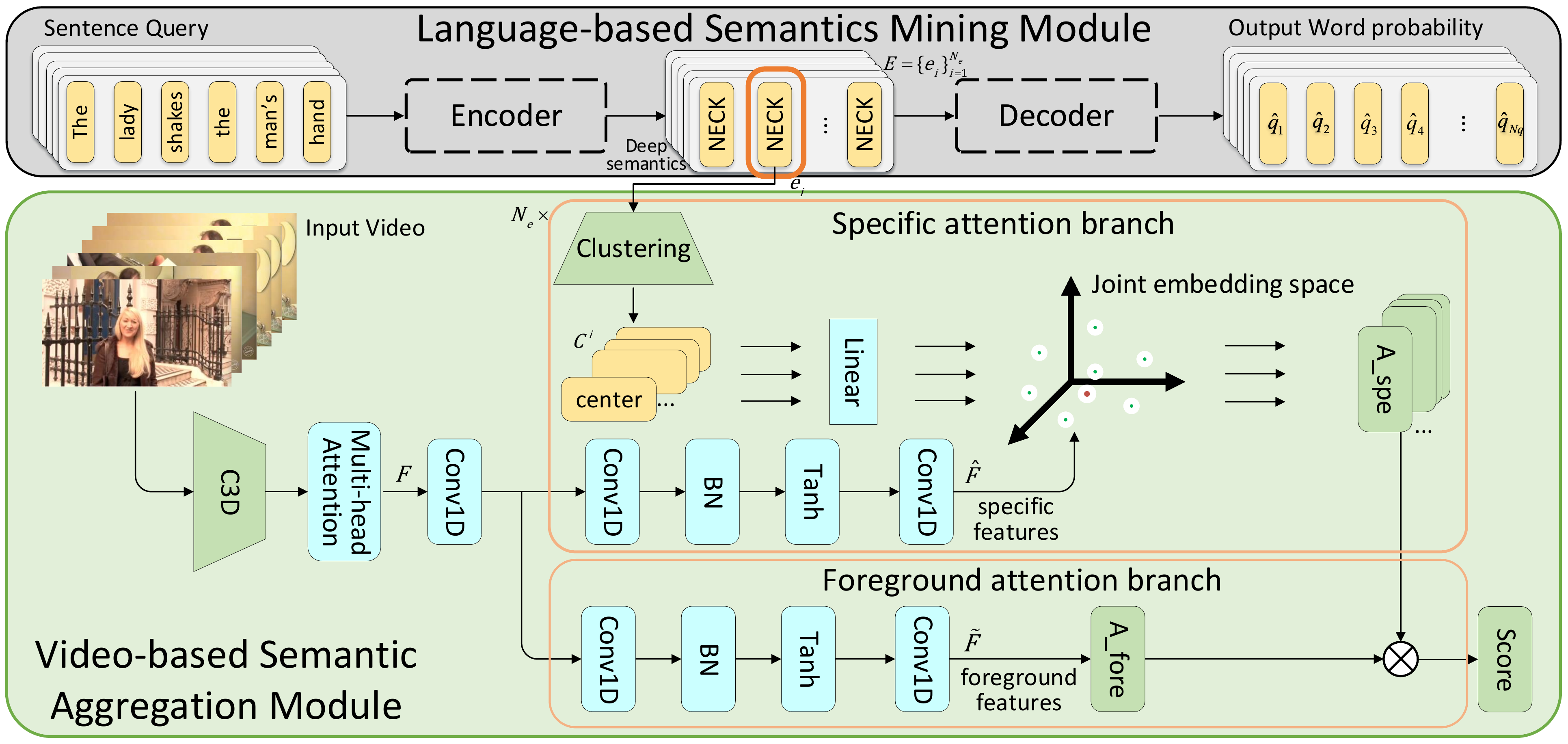}}
\end{center}
\vspace{-10pt}
\caption{The overall architecture of the proposed DSCNet for unsupervised TVG task. Given a query set, we first develop a language-based semantic mining module to learn the deep semantic for all queries by an encoder-decoder model. Then a video-based semantic aggregation module is proposed to compose the possible activities referring to the deep semantic clusters.}
\vspace{-10pt}
\label{fig:pipeline}
\end{figure*}

\noindent \textbf{Unsupervised Learning.}
Recently, unsupervised methods \cite{soomro2017unsupervised,laina2019towards,su2019deep,gong2020learning} receive increasing attention in multi-modal retrieval task. Laina \cite{laina2019towards} and Su \cite{su2019deep} embed both video and text into a shared latent space, then maximally reconstruct the joint-semantics relations. Soomro \cite{soomro2017unsupervised} and Gong \cite{gong2020learning} propose unsupervised action localization and transform the task into a unsupervised frame-wise classification problem with the pre-defined action categories. Different from these retrieval methods, the unsupervised TVG task requires fine-grained video-query alignment for better predicting accurate start-end timestamps. To learn deep semantics of sentence queries, we utilize widely used encoder-decoder architecture \cite{Alpher18,Alpher19,Alpher20,Alpher21,Alpher22} to learn its unsupervised representations, which consists of an encoder to extract feature representations and a decoder to reconstruct the input data from the representations. Then we compose the activity contents among the video guided by these learnt deep semantic features.

\section{Deep Semantics Clustering Network}
\subsection{Preliminaries}
In the TVG task, we are provided with a training set of untrimmed videos $\bm{V} = \{\bm{v}_i\}^N_{i=1}$ and sentence queries $\bm{Q} = \{\bm{q}_i\}^M_{i=1}$,
where $\bm{v}_i$ and $\bm{q}_i$ are the $i$-th video and query, $N$ and $M$ are their corresponding numbers. Since our method is under unsupervised setting, we drop all label information between $\bm{V}$ and $\bm{Q}$ including both their  correspondence and annotated segment boundaries. 
% To learn the grounding strategy, we first mine the deep semantic features of the whole query set, and then train our video model by composing the possible foreground activity contents in each video with these deep semantics.

The overall architecture of our proposed Deep Semantic Clustering Network (DSCNet) is shown in Figure \ref{fig:pipeline}. 
We first develop an independent encoder-decoder network to learn deep semantic features for the whole query set. In particular, we extract the hidden representations of each sentence as deep semantic called ``neck", and gather the necks of all queries into different semantic clusters by a clustering algorithm. Furthermore, we devise a video-based aggregation model with two attention branches: specific attention branch and foreground attention branch. Specifically, the specific attention branch is devised to match the frame-wise feature with each semantic cluster, and the foreground attention branch is developed to distinguish the foreground-background events. 
% In practice, we iterate such grounding procedure multiple times.
% More details will be described in the following sections.

\subsection{Language-based Semantic Mining}
% In previous fully- or weakly-supervised methods, the fine-grained semantic information of queries are usually learned with the help of the supervision information. However, similar approach cannot work in the unsupervised setting. 
In this section, 
% we make attempts to learn an independent language module by extracting the internal information of each query to reconstruct its main meaning.
we make an attempt to extract the internal information of each query by reconstructing its main meaning with an independent encoder-decoder network. 
% To this end, we develop an encoder-decoder network with a well-designed loss function to learn its hidden-state features as the deep semantics of the input queries.
% To learn the discriminative hidden features of input queries, also named as deep semantics ``neck", a well-designed loss function is utilized here.
Meanwhile, a well-designed loss function $\mathcal{L}_{w}$ in Eq. (3) is proposed to learn the discriminative hidden feature of input queries, which is also named as deep semantic “neck” thereafter.

For encoder, given a sentence query, we first employ off-the-shelf Glove model \cite{Alpher25} to obtain its word-level embedding $\bm{W}\in \mathbb{R}^{L \times {d_w}}$, where $L$ is the sentence length and ${d_w}$ is the embedding dimension. Then we feed $\bm{W}$ into a two-layer LSTM network and use the last hidden unit as the sentence-level representation $\bm{r}_e \in \mathbb{R}^{d_r}$.
To separate different latent semantic representations which describe different aspects in $\bm{r}_e$, we further feed the sentence-level representation $\bm{r}_e$ into multiple two-layer perceptrons to obtain the hidden features of the encoder-decoder model, called ``necks", which serve as the implicit semantics of queries.
Specifically, we denote the necks of a query as $\bm{E} = \{\bm{e}_i\}^{N_e}_{i=1} \in \mathbb{R}^{N_e \times d_e}$,
where ${N_e}$ is the neck number and $d_e$ is dimension.

To ensure that the learned necks have contained the most crucial information of the sentence, we further adopt a decoder module to reconstruct the original sentence with these necks.
In details, we first aggregate the information from all necks to output a new sentence-level representation $\bm{r}_o \in \mathbb{R}^{d_r}$ by other multiple two-layer perceptrons with further concatenation. Then, we feed it into another two-layer LSTM network and a linear layer to construct the word-level sentence, which is expected to be the same as the input one.
% the architecture of LSTM in decoder is revised to one to many.
% \emph{L} single LSTM layers are serially connected and the output of prior LSTM serves as the input of the next one.
Specifically, for the $i$-th output word, the decoder outputs its score vector $\{p_{i,j}\}^{N_w}_{j=1}$, where $N_w$ is the size of vocabulary in the whole query set, and $p_{i,j}$ means the probability distribution of $j$-th word in the vocabulary appearing at the $i$-th output word. Suppose the predicted probabilities of the word-level groundtruth location in original $L$-length input sentence as $\{p_{i,truth}\}_{i=1}^L$, we calculate Cross Entropy Loss $\mathcal{L}_{cel}$ with softmax function to supervise the output sentence as:
\begin{equation}
\mathcal{L}_{cel} = -\sum_{i=1}^L log( \displaystyle{\frac{exp(p_{i,truth})}{\sum_{j=1}^{N_w}exp(p_{i,j})}}).
\end{equation}
Besides, to ensure both the input and output sentences have the same sentence-level semantic meaning, we add a semantic loss $\mathcal{L}_{mse}$ calculated by the Mean Square Error function between both sentence-level representations $\bm{r}_e$ and $\bm{r}_o$.
More importantly, since we expect each neck in the query has unique semantic, 
we adopt a regularization term \cite{Alpher26} $\mathcal{L}_{dqa}$ to enforce necks $\{\bm{\bm{e}_i}\}_{i=1}^{N_e}$ be different from each other:
\begin{equation}
\mathcal{L}_{dqa} = \lVert (\bm{E}^{\top}\bm{E})- \lambda \bm{I} \rVert,
\end{equation}
where $\lVert \cdot \rVert$ denotes L2-norm, $\lambda \in (0,1]$ controls the extent of overlap between different necks. $\bm{I}$ is an identity matrix. By combining the above three losses with balanced parameters $\alpha_w$, $\beta_w$, we can adopt a multi-task loss function in the semantic mining module as follows:
\begin{equation}
\mathcal{L}_{w} = \mathcal{L}_{cel} + {\alpha_w}\mathcal{L}_{mse} + {\beta_w}\mathcal{L}_{dqa}.
\label{111}
\end{equation}

In a similar way, we can get the neck features for all sentences in $\bm{Q}$, where each sentence has $N_e$ necks.
Subsequently, for the $i$-th neck of all queries, we implement K-means clustering algorithm \cite{na2010research} to get $N_c$ centers upon them. Formally, the centers of the $i$-th necks are recorded as $\bm{C}^i = \{\bm{c}_j^i\}_{j=1}^{N_c} \in \mathbb{R}^{N_c \times d_e}$, $d_e$ is the dimension of each center feature. These centers can be regarded as discriminative semantic features. 
% We present visualization results of clustering in our experiment.
In the following video-based semantic aggregation module, such central semantic representations can be utilized for activity content composing.

\subsection{Video-based Semantic Aggregation}
To compose possible activities referring to the central semantic features from the queries, we develop a video-based semantic aggregation module which consists of a specific attention branch and a foreground attention branch.
During the training of the video module, we initialize a pseudo label for video frames as weak guidance to assist grounding, and utilize an iterative learning strategy to update and refine the pseudo labels for better training.

\noindent \textbf{Video feature encoding.}
Given a video, we first utilize a C3D \cite{Alpher27} network with a multi-head self-attention module \cite{vaswani2017attention} to extract the frame-wise features as $\bm{F} = \{\bm{f}_t\}^T_{t=1} \in \mathbb{R}^{T \times {d_v}}$, where $T$ is the number of frames in one video and $d_v$ is the channel dimension of frame-wise representation.
% Since the frame-wise feature can be matched with all semantic features, 

\noindent \textbf{Pseudo labels.} For specific semantic cluster $\bm{C}^i$, we initialize the pseudo labels $\bm{Y} = \{\bm{y}_j\}_{j=1}^{N_c} \in \mathbb{R}^{{N_c}\times T}$ to all $T$ frames, where each label denotes whether a specific frame $t$ matches the semantic cluster center $\bm{c}_j^i$. Specifically, we implement N-cut \cite{Alpher24} clustering with Gaussian kernel upon the concatenated feature $[\bm{f}_t;\bm{c}_j^i]$ to assign the binary label ${\bm{y}_j} \in \mathbb{R}^{T}$ label for each frame. 
Such clustering process can obtain coarse activity information. 
% In the next iterative learning process, the pseudo labels will be updated.
The fine-grained label would be obtained through the iterative learning process.

\noindent \textbf{Specific attention branch.}
In this branch, we aim to aggregate different semantics of the query set for better composing possible activities among the video.
After getting the semantic centers $\bm{C}^i=\{\bm{c}^i_j\}_{j=1}^{N_c}$ of $i$-th neck of the whole query set, we first project both language and video features into a joint embedding space, and denote their new features as $\hat{\bm{C}}^{i} \in \mathbb{R}^{{N_c} \times {d_{e'}}}$ and $\hat{\bm{F}} = \{\hat{\bm{f}}_t\}_{t=1}^T \in \mathbb{R}^{{T} \times {d_{e'}}}$, respectively. Then, we calculate the correlations between all frame-semantic pairs as the specific attention matrix $\bm{A}_{spe}$:
\begin{equation}
\bm{A}_{spe} = Softmax(\hat{\bm{C}}^i(\hat{\bm{F}})^{\top}) \in \mathbb{R}^{N_c \times T},
\end{equation}
where each row of $\bm{A}_{spe}$ denotes the similarities of all frames to a specific semantic center, and 
% the most matched frames will obtain higher scores and be composed into the activities.
those frames with the corresponding highest scores will be composed into the activities.

% To ensure that the composed activities containing the same semantic have similar representations and the positive-negative activity of the same video have larger distance, we attempt to design a loss function to yield a better video feature representation.
% Specifically, to make the best usage of the large video data, we take the above frame-level restriction to the higher video-level one. 
% Therefore, we aggregate the frame-wise semantic features guided by a specific semantic cluster to explore activity-specific common representation among videos.
In general, given a batch of training videos, we randomly sample $J$ semantic cluster centers and $Z$ videos.
For $v$-th video feature $\hat{\bm{F}}_v$, we can calculate its specific positive activity features guided by $j$-th semantic cluster center as:
\begin{equation}
\tilde{\bm{S}}_{v,j}^{p} = \bm{A}_{spe}[j,:] {\hat{\bm{F}}_v} \in \mathbb{R}^{1 \times d_{e'}}.
\end{equation}
We can also generate its negative specific feature by:
\begin{equation}
\bm{B}_{spe} = \frac{\bm{1}-\bm{A}_{spe}}{T},
\end{equation}
\begin{equation}
\tilde{\bm{S}}_{v,j}^{n} = \bm{B}_{spe}[j,:]{\hat{\bm{F}}_v} \in \mathbb{R}^{1 \times d_{e'}},
\end{equation}
where $\bm{1}$ here is a matrix with the same shape as $\bm{A}_{spe}$ filled by integer 1. 
Dividing by $T$ is for the purpose of normalization.
Since we expect the integrated positive features (different videos $v,u$ about the same semantic center $j$) having the same semantic information to be similar while the positive and negative features of the same video (video $v$) to be distinct,
our loss function $\mathcal{L}_{sab}$ of specific attention branch can be formulated as follows where $d(\cdot)$ is the cosine distance:
\begin{equation}
\mathcal{L}_{sim}^{(j,v)} =  \sum_{u=1,u\neq v}^{Z} max[d(\tilde{\bm{S}}_{v,j}^{p},\tilde{\bm{S}}_{u,j}^{p}) -\tau_1,0],
\end{equation}
\begin{equation}
\mathcal{L}_{dis}^{(j,v)} = \sum_{u=1,u\neq v}^{Z} max[d(\tilde{\bm{S}}_{v,j}^{p},\tilde{\bm{S}}_{u,j}^{p}) - d(\tilde{\bm{S}}_{v,j}^{p},\tilde{\bm{S}}_{v,j}^{n}) +\tau_2,0],
\end{equation}
\begin{equation}
\mathcal{L}_{sab} = \sum_{j=1}^{J} \sum_{v=1}^{Z} (\mathcal{L}_{sim}^{(j,v)} + \theta \mathcal{L}_{dis}^{(j,v)}),
\end{equation}
where $\tau_1$ and $\tau_2$ denote margins, $\theta$ is a weight coefficient.
In this way, the specific attention branch can aggregate most relevant frame-wise features corresponding to the specific semantic cluster center for activity composing.
% , leading to learn effective correlation between the frame-wise feature and semantic cluster feature for better activity composing and grounding.

\noindent \textbf{Foreground attention branch.}
Only using specific attention is not enough to provide accurate grounding, as it can not filter out all background activities composed by the irrelevant semantics features.
Therefore, as shown in Figure \ref{fig:pipeline}, we design a foreground attention branch with foreground features $\tilde{\bm{F}}$ and softmax attention output $\bm{A}_{fore} \in \mathbb{R}^{T\times 1}$, which can be obtained with several CNN layers. 
We first develop a triplet loss to distinguish the foreground-background frame-wise features according to the learned pseudo labels, where we aim to pull the frame representations of foreground (frames $u,v$) closer and push the frame representations of foreground-background (frames $u,o$) further in the feature space. Specifically, frame $v$ is the foreground frame which has the minimum distance to $u$, and frame $o$ is the background frame which has the maximum distance to $u$. The triple loss function can be formulated as:
\begin{equation}
\mathcal{L}_{trip} = \sum_{u =1}^Z max[d(\tilde{\bm{f}}_u^{p},\tilde{\bm{f}}_v^{p}) - d(\tilde{\bm{f}}_u^{p},\tilde{\bm{f}}_o^{b}) +\tau_3,0].
\end{equation}

% To further guide the grounding process of both specific attention matrix $\bm{A}_{spe}\in \mathbb{R}^{N_c \times T}$ and foreground attention matrix $\bm{A}_{fore}\in \mathbb{R}^{T \times 1}$ with pseudo labels $\bm{Y}\in \mathbb{R}^{T \times N_c}$, we calculate the cross entropy loss as below:
% \begin{equation}
% \mathcal{L}_{cls} = -\sum_{v=1}^{Z} \sum_{t=1}^T \bm{A}_{fore}[t,:] \cdot \mathcal{L}_{cls}^t,
% \end{equation}
% where $\bm{A}_{fore}[t,:]$ is utilized to give the weight to foreground-background frames, and $\mathcal{L}_{cls}^t$ denotes the grounding loss of each frame $t$ as:
% \begin{equation}
% h^t_j = Sigmoid(\bm{A}_{spe}[t,j]),
% \end{equation}
% \begin{equation}
% \mathcal{L}_{cls}^t = \sum_{j=1}^{N_c}[\bm{Y}[t,j] log(h^t_j) +(1-\bm{Y}[t,j])log(1-h^t_j)]
% \end{equation}
% where $\bm{A}_{spe}[t,j]$ means the similarity score from frame $t$ to semantic center $j$, and $h_j^t$ denotes whether the frame $t$ is the foreground referring to this semantic center.
% Combining the above three losses in two attention branches, we get the overall multi-task loss $\mathcal{L}_{v}$ in video grounding stage as: 
% \begin{equation}
% \mathcal{L}_{v} = \mathcal{L}_{cls} + {\alpha_v}\mathcal{L}_{sab} + \beta_v\mathcal{L}_{trip}.
% \label{222}
% \end{equation}
% where ${\alpha_v}$ and $\beta_v$ are hyper-parameters.

To predict the matching score of each frame by learning both specific attention matrix $\bm{A}_{spe}\in \mathbb{R}^{N_c \times T}$ and foreground attention matrix $\bm{A}_{fore}\in \mathbb{R}^{T \times 1}$ with the supervision of pseudo labels $\bm{Y}\in \mathbb{R}^{T \times N_c}$, we first calculate the score $h_j^t$ of $t$-th frame referring to semantic center $j$ and the foreground attention matrix $\bm{A}_{fore}$, and then formulate the grounding loss $\mathcal{L}_{cls}^t$ of each frame $t$ as:
\begin{equation}
h^t_j = \bm{A}_{spe}^{\top}[t,j] \times \bm{A}_{fore}[t,:],
\end{equation}
\begin{equation}
\mathcal{L}_{cls}^t = \sum_{j=1}^{N_c}[\bm{Y}[t,j] log(h^t_j) +(1-\bm{Y}[t,j])log(1-h^t_j)]
\end{equation}
where $\bm{A}_{spe}[t,j]$ denotes whether the frame $t$ is the positive frame referring to the semantic center $j$, and $\bm{A}_{fore}[t,:]$ is utilized to determine whether the frame is the foreground. Then, we calculate the grounding loss for whole frames as:
\begin{equation}
\mathcal{L}_{cls} = -\sum_{v=1}^{Z} \sum_{t=1}^T  \mathcal{L}_{cls}^t.
\end{equation}
Combining the aforementioned three losses in two attention branches, we get the overall multi-task loss $\mathcal{L}_{v}$ in the video grounding model as: 
\begin{equation}
\mathcal{L}_{v} = \mathcal{L}_{cls} + {\alpha_v}\mathcal{L}_{sab} + \beta_v\mathcal{L}_{trip}.
\label{222}
\end{equation}
where ${\alpha_v}$ and $\beta_v$ are hyper-parameters.

\begin{algorithm}[t!] 
\caption{Iterative learning process of video module} 
\label{alg::conjugateGradient} 
{\bf Input:} All semantic cluster centers $\bm{C}$ of the whole query set; video feature $\bm{F}$.
\begin{algorithmic}[1]
\STATE Init pseudo label based on $\bm{C}$ and $\bm{F}$
\STATE {\bf for} iteration $l \leftarrow 1$ to $L$ {\bf do}
\STATE \qquad {\bf for} neck $i \leftarrow 1$ to $N_e$ {\bf do}
\STATE \qquad \qquad Execute specific attention branch with $\emph{$\bm{C}^i$} = \{\bm{c}_j^i\}^{N_c}_{j=1}$ to obtain $\emph{$\hat{\bm{F}}_{v}$} = \{\hat{\bm{f}}^t_{v}\}^T_{t=1}$ and \emph{$\bm{A}_{spe}$};
\STATE \qquad \qquad Execute foreground attention branch to obtain \emph{$\bm{A}_{fore}$};
\STATE \qquad \qquad Generate the training samples by pseudo labels, and calculate the overall loss $\mathcal{L}_v$ for back-propagation;
\STATE \qquad \qquad Generate the new feature \emph{$\hat{\bm{F}}_{v}$}, and utilize it to update the pseudo labels;
\STATE \qquad {\bf end}
\STATE {\bf end}
\end{algorithmic}
\end{algorithm}

\noindent \textbf{Iterative learning.}
We use an iterative optimization strategy to train our video module. In each iteration: (1) we update the pseudo label on each frame by applying the cluster algorithm on the new feature $[\hat{\bm{f}}_t;\bm{c}_j^i]$, where $\hat{\bm{f}}_t$ is the learned feature in the specific branch as it contains more semantic-aware contexts. (2) we calculate and back-propagate the loss $\mathcal{L}_{v}$ for updating the video model. The overall training process is shown in Algorithm \ref{alg::conjugateGradient}. During the iterative training,  the grounding module gradually finds the important frames of the video and yields a better frame-wise feature representation. Such precise feature representations can further lead to more precise pseudo labels obtained from the clustering process, and in turn provides better supervisions for the grounding. We show the effectiveness of the iterative learning process in our experiments.

\subsection{Inference}
When testing, we directly utilize the generated $N_e$ necks feature $\bm{E}$ of the input query as semantic cluster centers, and feed them into the video module to match with frame-wise features for generating corresponding specific attention $\bm{A}_{spe} \in \mathbb{R}^{N_e \times T}$ and foreground attention $\bm{A}_{fore} \in \mathbb{R}^{1 \times T}$ for activity composing. Specifically, we element-wisely multiply $\bm{A}_{spe}$ and $\bm{A}_{fore}$ for each neck, and feed the results of all necks to softmax layers with a further element-wise multiplication. The final attention scores of size $1 \times T$ denotes a joint probability of all semantics. Finally, we locate the frame with the highest score as the basic predicted segment, and add the left/right frames into the segment if the ratio of their scores to the frame score of the closest segment boundary is less than a threshold. We repeat this step until no frame can be added.
% generate the segment using the frame of the highest score and iteratively merge the nearby frames if their scores are large than a threshold.

\section{Experimental Results}
\subsection{Datasets and Evaluation}
\noindent \textbf{Charades-STA.}
This dataset is built from the Charades \cite{Alpher28}
dataset and transformed into video temporal grounding task by \cite{Alpher06}. It contains 16128 video-sentence pairs with 12408 pairs used for training and 3720 for testing. The videos are about 30 seconds on average. The annotations are generated by sentence decomposition and keyword matching with manually check.

\noindent \textbf{ActivityNet Captions.}
This dataset is built from
ActivityNet v1.3 dataset \cite{Alpher29} for dense video captioning. It contains 20000 YouTube videos with 100000 queries. We follow the public split of the dataset that contains a training set and two validation sets val 1 and val 2. On average, videos are about 120 seconds and queries are about 13.5 words. 

\begin{table*}
\small
\begin{center}
\scalebox{0.95}{
\begin{tabular}{l|c|cccccc|cccc}
\hline
\multirow{3}*{Methods} & \multirow{3}*{Mode} & \multicolumn{6}{c|}{Charades-STA} & \multicolumn{4}{c}{ActivityNet Captions} \\ \cline{3-12}
~ & ~ & R@1 & R@1 & R@1 & R@5 & R@5 & R@5 & R@1 & R@1 & R@5 & R@5\\
~ & ~ & IoU=0.3 & IoU=0.5 & IoU=0.7 & IoU=0.3& IoU=0.5 & IoU=0.7 & IoU=0.3 & IoU=0.5 & IoU=0.3 & IoU=0.5\\
\hline\hline
Rondom & FS & 20.12 & 8.51 & 3.03 & 68.42 & 37.12 & 14.06 & 18.64 & 7.73 & 52.78 & 29.49\\
VSA-RNN & FS & - & 10.50 & 4.32 & - & 48.43 & 20.21 & 39.28 & 24.43 & 70.84 & 55.52\\
VSA-STV & FS & - & 16.91 & 5.81 & - & 53.89 & 23.58 & 41.71 & 24.01 & 71.05 & 56.62\\
CTRL & FS & - & 23.62 & 8.89 & - & 58.92 & 29.52 & - & - & - & -\\
TGN & FS & - & - & - & - & - & - & 43.81 & 27.93 & 54.56 & 44.20 \\
EFRC & FS & 53.00 &33.80 & 15.00 & 94.60 & 77.30 & 43.90 & - & - & - & -\\
2D-TAN & FS & - & 39.81 & 23.25 & - & 79.33 & 52.15 & 59.45 & 44.51 & 85.53 & 77.13 \\
DRN & FS & - & 45.40 & 26.40 & - & 88.01 & 55.38 & - & 45.45 & - & 77.97\\
\hline\hline
TGA & WS & 32.14 & 19.94 & 8.84 & 86.58 & 65.52 & 33.51 & - & - & - & -\\
SCN & WS & 42.96 & 23.58 & 9.97 & 95.56 & 71.80 & 38.87 & 47.23 & 29.22 & 71.45 & 55.69\\
CTF & WS & 39.80 & 27.30 & 12.90 & - & - & - & 44.30 & 23.60 & - & -\\
MARN & WS & 48.55 & 31.94 & 14.18 & 90.70 & 70.00 & 37.40 & 47.01 & 29.95 & 72.02 & 57.49\\
VGN & WS & - & 32.21 & 15.68 & - & 73.50 & 41.87 & 50.12 & 31.07 & 77.36 & 61.29\\
\hline\hline
\textbf{DSCNet} & US & \textbf{44.15} & \textbf{28.73} & \textbf{14.67} & \textbf{91.48} & \textbf{70.68} & \textbf{35.19} & \textbf{47.29} & \textbf{28.16} & \textbf{72.51} & \textbf{57.24} \\
\hline
\end{tabular}}
\end{center}
\vspace{-10pt}
\caption{Performance comparisons for video grounding on both Charades-STA and ActivityNet Captions datasets, where FS: fully-supervised setting, WS: weakly-supervised setting and US: unsupervised setting.}
\label{tab:Cha}
% \vspace{-4pt}
\end{table*}

\begin{table*}
\begin{center}
\small
\setlength{\tabcolsep}{1.7mm}{
\begin{tabular}{l|ccc|ccc|cccccc}
\hline
\multirow{2}*{Method} & \multicolumn{3}{c|}{Language Module} & \multicolumn{3}{c|}{Video Module} & R@1 & R@1 & R@1 & R@5 & R@5 & R@5 \\ ~ &
$\mathcal{L}_{cel}$ & $\mathcal{L}_{dqa}$ & $\mathcal{L}_{mse}$ & $\mathcal{L}_{sab}$ &  $\mathcal{L}_{trip}$ & $\mathcal{L}_{cls}$ & IoU=0.3 & IoU=0.5 & IoU=0.7 & IoU=0.3 & IoU=0.5 & IoU=0.7\\
\hline\hline
baseline & $\checkmark$ & $\times$ & $\times$ & $\times$ & $\times$ & $\checkmark$ & 28.38 & 16.12 & 7.86 & 76.18 & 48.65 & 20.21\\
+ $\mathcal{L}_{dqa}$ & $\checkmark$ & $\checkmark$ & $\times$ & $\times$ & $\times$ & $\checkmark$ & 33.54 & 20.07 & 9.09 & 80.16 & 55.41 & 23.72 \\
+ $\mathcal{L}_{w}$& $\checkmark$ & $\checkmark$ & $\checkmark$ & $\times$ & $\times$ & $\checkmark$ & 36.78 & 23.14 & 10.65 & 82.98 & 60.29 & 26.31 \\
+ $\mathcal{L}_{w}$ + $\mathcal{L}_{trip}$+ $\mathcal{L}_{cls}$& $\checkmark$ & $\checkmark$ & $\checkmark$ & $\times$ & $\checkmark$ & $\checkmark$ & 40.23 & 25.32 & 12.11 & 86.83 & 65.74 & 30.45 \\
+ $\mathcal{L}_{w}$ + $\mathcal{L}_{sab}$+ $\mathcal{L}_{cls}$& $\checkmark$ & $\checkmark$ & $\checkmark$ & $\checkmark$ & $\times$ & $\checkmark$ & 41.97 & 26.01 & 12.39 & 87.61 & 67.33 & 32.60 \\ \hline
+ $\mathcal{L}_{w}$ + $\mathcal{L}_{v}$ & $\checkmark$ & $\checkmark$ & $\checkmark$ & $\checkmark$ & $\checkmark$ & $\checkmark$ & \textbf{44.15} & \textbf{28.73} & \textbf{14.67} & \textbf{91.48} & \textbf{70.68} & \textbf{35.19} \\ \hline
\end{tabular}}
\end{center}
\vspace{-10pt}
\caption{The analysis of each component in DSCNet, via ablation study on Charades-STA.}
\label{tab:mainab}
\vspace{-10pt}
\end{table*}

\noindent \textbf{Evaluation.}
Following prior work \cite{Alpher06}, we adopt ``R@$N$, IoU=$\theta$'' as our evaluation metrics, which is defined as the percentage of at least one of top-$N$ selected moments having IoU scores larger than $\theta$.

\subsection{Implementation Details}
In order to make a fair comparison with previous works, we utilize C3D to extract video features and Glove to obtain word embeddings. 
As some videos are too long, we set the length of video feature sequences to 128 for Charades-STA and 256 for ActivityNet Captions. We fix the query length to 10 in Charades-STA and 20 in ActivityNet Captions. We set neck number $N_e$ to 4 for Charades-STA and 8 for ActivityNet Captions, and set cluster number $N_c$ to 16.
The LSTM Layers in language encoder and decoder are both 2 layers architecture with 512 hidden size. 
The dimension of joint embedding space \emph{$d_{e'}$} is set to 1024. We utilize Adam optimizer with the initial learning rate as 0.0001 for language module and 0.0005 for video module.
The hyper-parameters $\theta, \tau_1, \tau_2, \tau_3$ are set as 1.0, 0.0001, 0.0001, 0.5. $\lambda$ is set to 0.5.
And $\alpha_w, \beta_w, \alpha_v, \beta_v$ in Eq. (\ref{111}) and (\ref{222}) are all set as 0.5. The inference threshold is set to 0.8 in ActivityNet and 0.9 in Charades-STA.

\subsection{Comparisons with state-of-the-arts}
\noindent \textbf{Comparison on Charades-STA.} 
We first compare our model DSCNet with the state-of-the-art methods on Charades-STA dataset, shown in Table \ref{tab:Cha}. Specifically, for metrics R@1, IoU$\in$\{0.3,0.5,0.7\}, the results achieved by our method in the unsupervised setting (US) are comparable to the results obtained by the state-of-the-art fully-supervised (FS) and weakly-supervised (WS) methods. For R@5, we also have similar observations. 

\noindent \textbf{Comparison on ActivityNet Captions.} We also presents the results on ActivityNet Captions, shown in Table \ref{tab:Cha}. We compare our method DSCNet with other recent state-of-the-art FS and WS video grounding methods. Even without any video-query annotations, our method is able to achieve 47.29\%, 28.16\%, 72.51\%, 57.24\% in all metrics, showing competitive performance comparing to almost all weakly supervised methods.

\subsection{Ablation Study}
In this section, we conduct ablation study to validate the effectiveness of each components in our methods. All experiments are conducted on Charades-STA dataset.

\noindent \textbf{Effect of each component.}
To analyze how each model component contributes to the task, we perform ablation study as shown in Table \ref{tab:mainab}. We use the model in which the language-based encoder-decoder is only combined with reconstruction loss $\mathcal{L}_{cel}$, and the video module is combined with grounding loss $\mathcal{L}_{cls}$ as our baseline. Then we add the regularization loss $\mathcal{L}_{dqa}$ to the baseline model and improve R@1 IoU=0.3 from 28.38\% to 33.54\%, R@1 IoU=0.5 from 16.12\% to 20.07\%, demonstrating the importance of learning different semantic features. The sentence-level semantic loss $\mathcal{L}_{mse}$ can further improve the hidden features learning of the auto-encoder model. The special attention branch loss $\mathcal{L}_{sab}$ and triple loss $\mathcal{L}_{trip}$ also bring significant improvements by yielding better frame-wise representations. 
From the table, we can see that jointly combining all the loss functions achieves the superior overall performance.

\noindent \textbf{How to select the cluster center.} As shown in Table \ref{tab:language}, we investigate how the strategy of selecting the center \emph{$\bm{C}$} affects the grounding results. We can find that randomly selection without clustering performs poorly, since it lacks sufficient semantics to compose all possible activities. The ``cluster center" (we directly choose the averaged cluster center) performs much better than the ``cluster sample", where we randomly choose one sample in each cluster. The reason could be the central embedding in each cluster contains the most representative semantic to cover other samples.

\begin{table}
\begin{center}
\small
\begin{tabular}{l|c|c|c|c}
\hline
Methods & \tabincell{c}{R@1\\IoU=0.3}&\tabincell{c}{R@1\\IoU=0.5}&\tabincell{c}{R@5\\IoU=0.3}& \tabincell{c}{R@5\\IoU=0.5}\\
\hline\hline
random & 23.79 & 11.24 & 73.61 & 42.58 \\
cluster sample & 39.67 & 23.96 & 86.13 & 66.49 \\
cluster center & \textbf{44.15} & \textbf{28.73}  & \textbf{91.48} & \textbf{70.68}\\
\hline
\end{tabular}
\end{center}
\vspace{-10pt}
\caption{Ablation study on the selection of the semantic cluster center $\bm{C}$.}
\label{tab:language}
\vspace{-6pt}
\end{table}

\begin{table}
\begin{center}
\small
\begin{tabular}{l|c|c|c|c}
\hline
Methods & \tabincell{c}{R@1\\IoU=0.3}&\tabincell{c}{R@1\\IoU=0.5}&\tabincell{c}{R@5\\IoU=0.3}& \tabincell{c}{R@5\\IoU=0.5}\\
\hline\hline
2 necks & 33.06 & 19.70 & 79.39 & 53.53 \\
4 necks & \textbf{44.15} & \textbf{28.73} & \textbf{91.48} & \textbf{70.68} \\
8 necks & 40.90 & 25.03 & 87.56 & 65.40 \\
\hline\hline
1 iteration & 29.38 & 15.61 & 81.13 & 60.08 \\
3 iteration & 39.98 & 24.87  & 88.09 & 66.77 \\
5 iteration & \textbf{44.15} & \textbf{28.73} & 91.48 & \textbf{70.68} \\
7 iteration & 44.06 & 28.39 & \textbf{91.82} & 70.41 \\
\hline
\end{tabular}
\end{center}
\vspace{-10pt}
\caption{Ablation study on the neck number and iteration number on Charades-STA dataset.}
\label{tab:neck}
\vspace{-10pt}
\end{table}

\noindent \textbf{Comparing different neck numbers.}
For the number of necks, as shown in Table \ref{tab:neck}, using 2 necks which contains complex information leads to poor scores. Using 4 necks achieves the best performance while using 8 necks is slightly lower. Therefore, we choose 4 as the neck number on Charades-STA dataset in our experiments.

\noindent \textbf{Comparing different iterative learning times.}
Table \ref{tab:neck} also show the performance on different iterations. As the number of iterations increases, the performance becomes better. Our model achieves the best results with 5 iterations. We do not see more improvements by increasing iterations after 5.

\subsection{Visualization Results}
\begin{figure}[t!]
% \vspace{-10pt}
\begin{center}
{\includegraphics[width=0.48\textwidth]{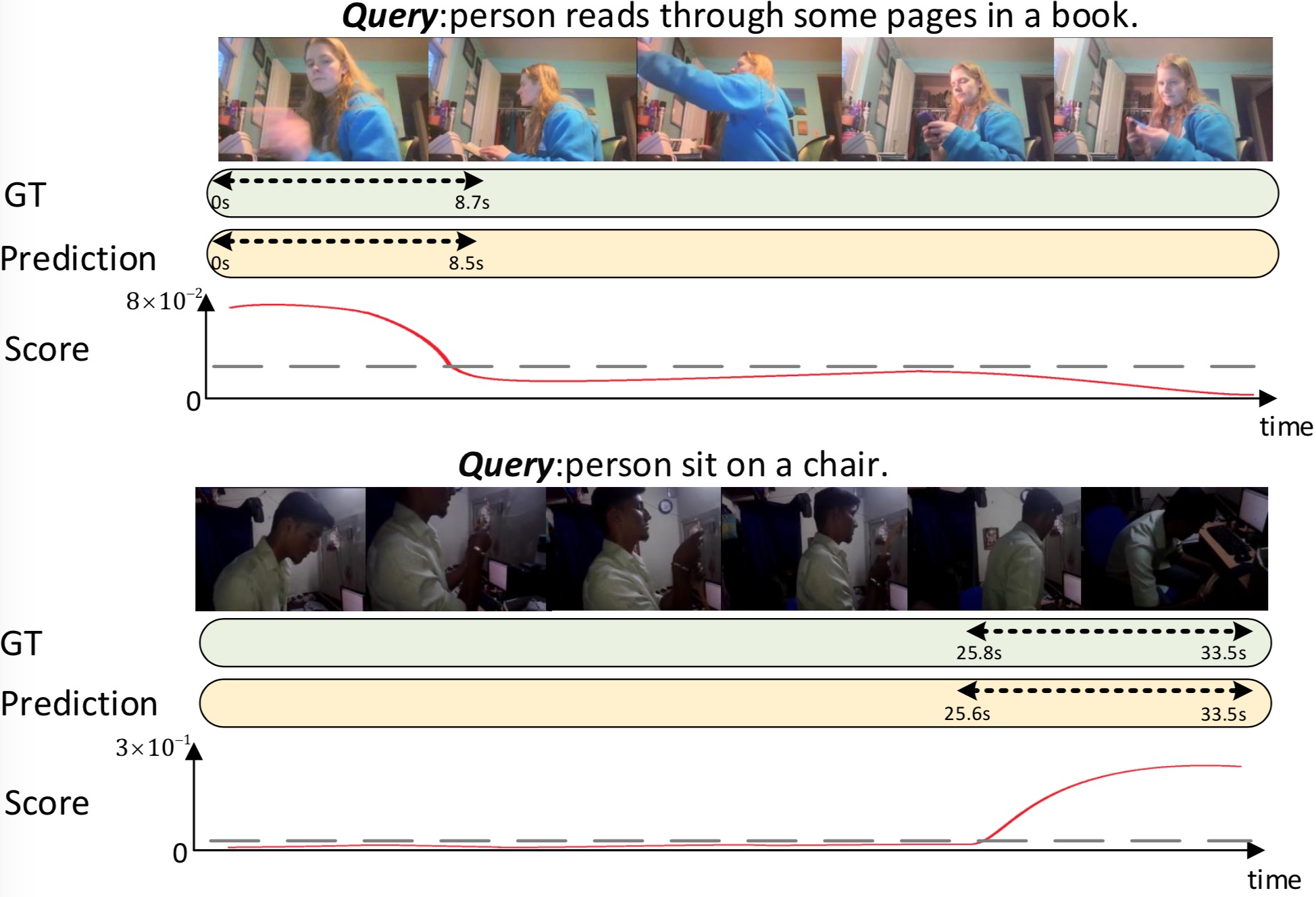}}
\end{center}
\vspace{-10pt}
\caption{Qualitative results on charades-STA dataset. ``GT" is the annotation of the ground-truth segment and ``Prediction" is our grounding result. The score value in the red curve of each video denotes the probability of each frame.}
\label{fig:gtandpre}
% \vspace{-10pt}
\end{figure}

\begin{figure}[t!]
\begin{center}
{\includegraphics[width=0.48\textwidth]{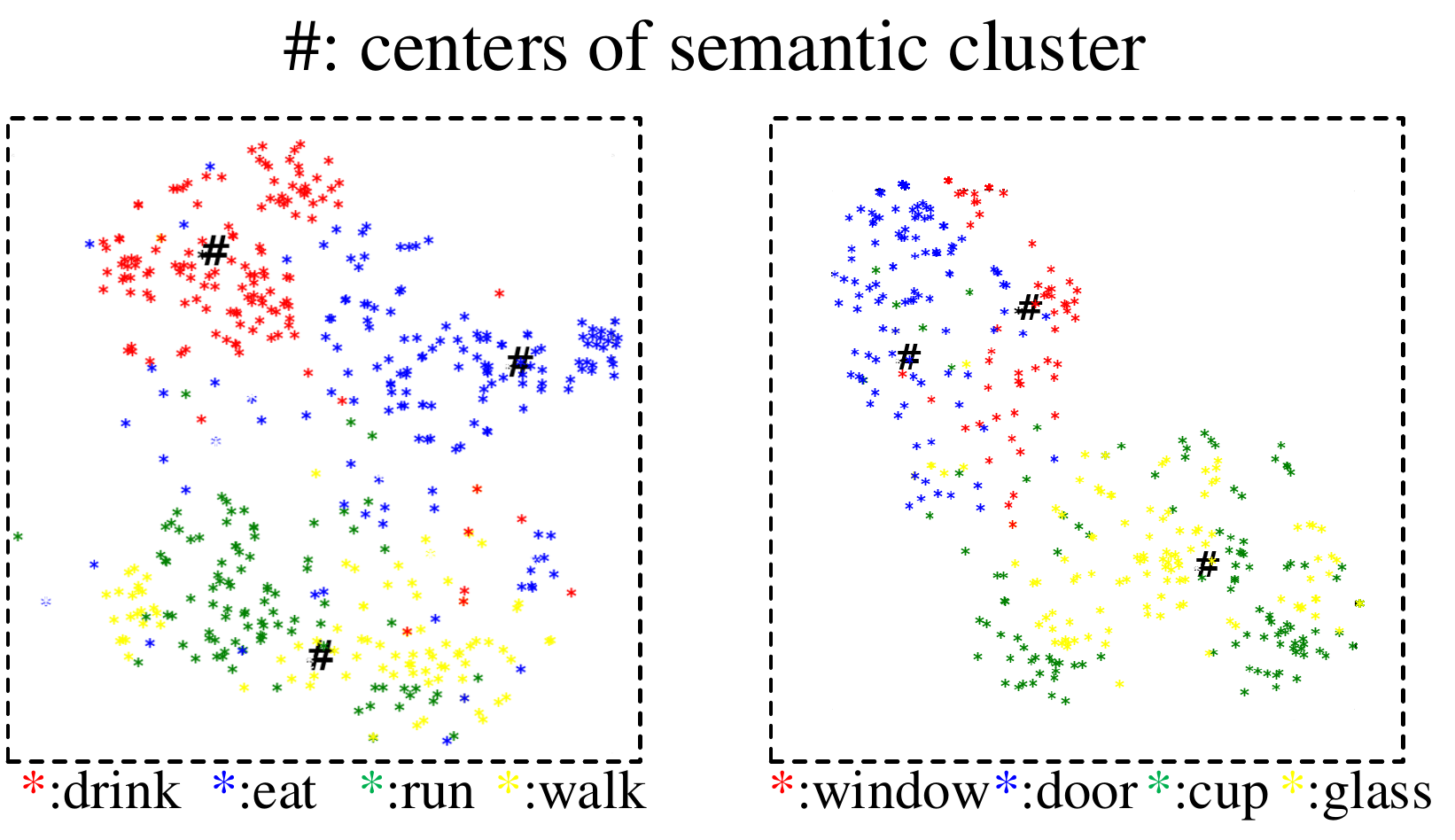}}
\end{center}
\vspace{-10pt}
\caption{Example of the semantic clustering results of the language-based semantic mining module.}
\label{fig:cluster}
\vspace{-10pt}
\end{figure}

Figure \ref{fig:gtandpre} shows some grounding results of our DSCNet on Charades-STA dataset. Figure \ref{fig:cluster} shows the semantic clustering results of the language module, 
in which we partially visualize the clusters related to actions and objects by utilizing T-SNE \cite{Alpher37}. We select sentences containing specific words from the test set of Charades-STA for visualization.  
For class actions, we take ``drink'', ``eat'', ``run'', ``walk'' as the examples. Through clustering, the actions ``drink'', ``eat'' are quite different from actions ``run'', ``walk'' since they generally appear in different scenarios (indoor vs. outdoor). Meanwhile, we can observe that there is a distinct margin between the ``drink'', ``eat'' and ``run'', ``walk'' as shown in the left figure. 
Furthermore, we can also find that actions ``drink'' and ``eat'' are well separated while ``run'' and ``walk'' are not, this is because: 
% (1) the surrounding objects are different when actions ``drink'' and ``eat'' appear, such as ``drink" with ``cup", ``eat" with ``plate"; (2) 
actions ``run'' and ``walk'' are quite close in semantics, and can be substituted for each other in some circumstances.
The right figure shows the clustering results on multiple objects. It illustrates that the objects ``cup", ``glass" are the intra-pairs, and the objects ``window", ``door" are the inter-pairs.

\section{Conclusion}
In this paper, we propose a novel Deep Semantic Clustering Network (DSCNet), to solve the temporal video grounding (TVG) task under the unsupervised setting. We first mine deep semantic features from all sentences and apply clustering on them to obtain the universal textual representation of the whole query set. Then, we compose the possible activities among the videos guided by the extracted deep semantic features. Specifically, we design two attention branches with the novel loss function for grounding. 
Our method is evaluated on two benchmark datasets and achieves decent performances, compared with most fully/weekly supervised baselines. The future work includes applying DSCNet to other tasks/datasets \cite{li2020hero,lei2020tvr}, and leveraging local/global features to learn better video-text representations. Following the idea of DSCNet, we would like to explore how to use more unannotated data in supervised manner. 

% \clearpage
\section{Acknowledgements} 
This work was supported in part by the National Natural Science Foundation of China (NSFC) under Grant (No.61972448, No.62172068, No.61802048).

\bibliography{egbib.bib}
\end{document}